\documentclass[sigconf]{acmart}

\usepackage{booktabs} % For formal tables

\usepackage[margin=0.5cm]{subcaption}
\usepackage{tabularx}
\usepackage{amsmath}
\usepackage{bm}
\renewcommand\footnotetextcopyrightpermission[1]{} % removes footnote with conference information in first column
\acmConference[CIKM2019]{The 28th ACM International Conference on Information and Knowledge Management}{November 3rd-7th, 2019}{Beijing, China}
\acmYear{2019}
\copyrightyear{2019}

\acmArticle{4}
\acmPrice{15.00}

\usepackage{xcolor}
\newcommand {\out}[1]{}

\begin{document}
\title{Adapting Visual Question Answering Models for Enhancing Multimodal Community Q\&A Platforms}
% \titlenote{Produces the permission block, and
%   copyright information}
% \subtitle{Extended Abstract}
% \subtitlenote{The full version of the author's guide is available as
%   \texttt{acmart.pdf} document}

% \author{Avikalp Srivastava\\
% Carnegie Mellon University\\
% avikalps@cs.cmu.edu\\
% \And
% Hsin-Wen Liu\\
% Waseda University\\
% stephanie1125@toki.waseda.jp
% \And
% Sumio Fujita\\
% Yahoo Japan Corporation\\
% sufujita@yahoo-corp.jp
% }

\author{Avikalp Srivastava}
\affiliation{%
  \institution{Carnegie Mellon University}
}
\email{avikalps@cs.cmu.edu}

\author{Hsin-Wen Liu}
\affiliation{%
  \institution{Waseda University}
}
\email{stephanie1125@toki.waseda.jp}

\author{Sumio Fujita}
\affiliation{%
  \institution{Yahoo Japan Corporation}
}
\email{sufujita@yahoo-corp.jp}

\begin{abstract}
Question categorization and expert retrieval methods have been crucial for information organization and accessibility in community question \& answering (CQA) platforms. Research in this area, however, has dealt with only the text modality. With the increasing multimodal nature of web content, we focus on extending these methods for CQA questions accompanied by images. Specifically, we leverage the success of representation learning for text and images in the visual question answering (VQA) domain, and adapt the underlying concept and architecture for automated category classification and expert retrieval on image-based questions posted on Yahoo! Chiebukuro, the Japanese counterpart of Yahoo! Answers. 

To the best of our knowledge, this is the first work to tackle the multimodality challenge in CQA, and to adapt VQA models for tasks on a more ecologically valid source of visual questions. Our analysis of the differences between visual QA and community QA data drives our proposal of novel augmentations of an attention method tailored for CQA, and use of auxiliary tasks for learning better grounding features. Our final model markedly outperforms the text-only and VQA model baselines for both tasks of classification and expert retrieval on real-world multimodal CQA data.

\end{abstract}

\maketitle

\section{Introduction}

Community question \& answering (CQA) platforms enable users to crowd-source answers to posted queries, search and explore questions, and share knowledge through answers. 
%As the size of the user base increases, so does the information content, making it imperative to carefully design methods for categorizing and organizing information and identifying relevant content for personalized recommendations. 
% Reviewer1 comment
As the number of users increases, so does the information content; this makes it imperative to carefully design methods for categorizing and organizing information and identifying relevant content for personalized recommendations.
Such end-tasks are of significant practical importance to CQA platforms, making them a big focus in information retrieval and natural language processing domains.

The CQA task of automatic question classification is useful for tagging newly posted questions and suggesting an appropriate question category to the asking user. It plays a crucial role in enabling users to find and answer questions in their area of expertise, thereby also facilitating effective answering.
% and also acts as a good first step towards potentially answering the question. 
% This has led to many studies focusing on this kind of classification \cite{zhang2003question}, \cite{saha2013discriminative}, \cite{stanley2013predicting}, \cite{tamaki2018classifying}.
Another useful problem to solve is that of retrieving ``experts''. Here, the aim is to identify and retrieve users from the community who are likely to provide answers to a given question. This provides an efficient way to make the community well-knit, provide better content to askers, and recommend only the relevant questions from a gigantic pool of queries to the potential answerers. 
% Previous works include \cite{zheng2017deep}, \cite{zhao2016expert}, \cite{riahi2012finding}, \cite{li2011question}, \cite{liu2005finding} among others.

\begin{figure*}[h]
\centering
\begin{subfigure}{.42\textwidth}
  \centering
  \includegraphics[width=\linewidth]{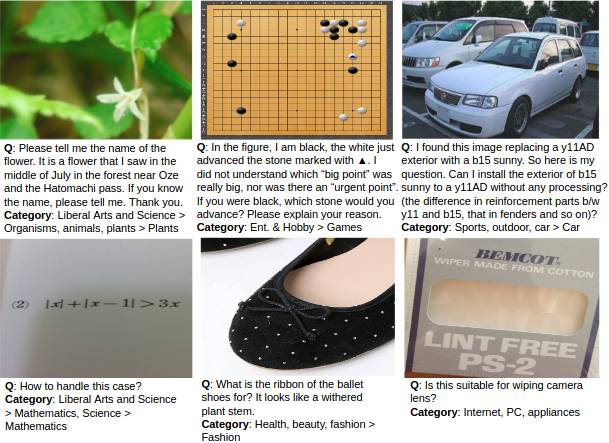}
  \label{fig:sub1}
\end{subfigure}%
\begin{subfigure}{.43\textwidth}
  \centering
  \includegraphics[width=\linewidth]{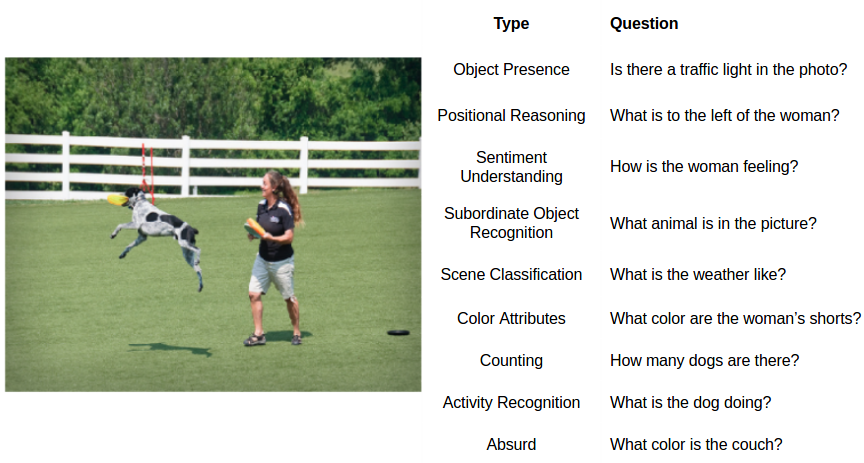}
  \label{fig:sub2}
\end{subfigure}
\caption{
Question image-text pairs sampled from (a) YC-CQA site (translated); (b): VQA, taken from \cite{kafle2017analysis}.}
\label{fig:1}
\end{figure*}

A recurring feature in these tasks has been that the data is comprised only of text. Research datasets from Stack Exchange, Quora, and collections like TREC-QA rarely contain questions with a combination of text and images. In this work, we tackle data from the CQA website Yahoo! Chiebukuro (YC-CQA), where questions accompanied by an image form a considerable percentage (${\sim}10\%$) of the total posted questions (Figure \ref{fig:1}(a)). With Stack Exchange sites supporting images (${\sim}7$\%, 11\%,  12\% and 20\% image-based questions on computer science, data science, movies, and anime stackexhange sites respectively), not to mention the numerous image-based threads on discussion platforms like Reddit, the advantages of our solutions for multimodal CQA are not limited to Chiebukuro.

Models using only text can give reasonable performances for multimodal CQA tasks (as we will see in our results), but there is potential to gain substantial improvements by utilizing the image data. 
% While many CQA samples have images that contain very little information, are irrelevant, or have characteristics that are rarely repeated across samples (as in the bottom-right example in Figure \ref{fig:1}(a)),
It is easy to identify a couple of broad categories where image data will be essential for our end-tasks: i) where the image contains the actual question, and the question loses meaning without the image (Figure \ref{fig:1}(a) bottom-left example), and ii) where the image is necessary to make sense of the question text (top-mid \& top-right examples in Figure \ref{fig:1}(a)). Images can also help reinforce the inferences from textual features (Figure \ref{fig:1}(a) top-left), or provide disambiguation over multiple topics inferred from text (`plants' and `shoes' in Figure \ref{fig:1}(a) bottom-mid example). 

Therefore, we focus on methods to best exploit the combined image-text information from multimodal CQA questions. 
% Our main tasks are classification and expert retrieval. 
Considering existing research at the intersection of vision and text, visual question answering models are dependent on deriving rich representations that encode a combined understanding of question's text-image pair. 
% and use a final classification layer to get the answer.
Thus, to not reinvent the wheel, we leverage the success of VQA architectures in deriving such joint representations, and build novel augmentations to adapt them for CQA tasks.
% for our tasks on community Q\&A. 

% An additional motive is to enable our developed models to be used to answer external knowledge based factoid questions (similar to \cite{wang2015explicit} \& \cite{yeh2008photo}) on CQA. 
% With this in mind, and to not reinvent the wheel, we leverage the success of visual question answering (VQA) models in dealing with text-image question pairs, and build novel augmentations to adapt them for CQA tasks. 

% \begin{figure*}[ht]
% \centering

% %\begin{subfigure}{.49\textwidth}
% \begin{subfigure}{.65\textwidth}
%   \centering
%   \includegraphics[width=\linewidth]{cqa_examples}
%   \label{fig:sub1}
%   \caption{Question image-text pairs sampled from Yahoo! JAPAN Chiebukuro community Q\&A site (translated)}
% \end{subfigure}\\\vspace{0.2cm}
% %\begin{subfigure}{.49\textwidth}
% \begin{subfigure}{.65\textwidth}
%   \centering
%   \includegraphics[width=\linewidth]{vqa_examples}
%   \label{fig:sub2}
%   \caption{Image-text pairs for VQA, taken from \cite{kafle2017analysis}.}
% \end{subfigure}
% \caption{Typical question image-text pairs from (a) CQA and (b) VQA}
% %\caption{(a): Question image-text pairs sampled from Yahoo! Chiebukuro community Q\&A site (translated); (b): Typical image-text pairs for VQA, taken from \cite{kafle2017analysis}.}
% \label{fig:1}
% \end{figure*}

In its most common form, the VQA task \cite{antol2015vqa} is modeled as a classification task involving an image-question pair (Figure \ref{fig:1}(b)) and selecting an answer from a fixed set of top possible answers.
% outputting an answer either as a small generated sequence, or using classification on a predefined set.
The main ideas behind its proposal has been to connect the advances in computer vision and NLP, so as to provide an ``AI-complete" task.
However, given the nature of the questions and images, its direct practical applicability is limited. The questions are short, direct, and query the image, or at the most require common sense or objective encyclopedic knowledge. This is in contrast with the nature of questions found on 
the web
% discussion and community QA platforms, 
where askers seek human expertise, and the question texts provide context outside the input image, or are supported by the image.
It is therefore important to properly identify and resolve the shortcomings of VQA models to enable better understanding of the image-text data from CQA.

While our contributions can be viewed under a more general lens, it is worth noting that given the significant percentage that image-based questions occupy on Chiebukuro, and the current policy on the site making it mandatory for asking users to provide a category from among hundreds of choices, improving automated category classification simplifies the introduction of the feature that suggests appropriate category to the askers. It can even allow them to skip this part by assigning the predicted category automatically. Providing better expert retrieval has the obvious benefit of improving the responsiveness and quality of the QA service as a whole. More generally, the decision to use VQA-inspired end-to-end learning architectures makes our models generalizable and usable for image-based sections on other QA/discussion platforms, along with possessing the potential for  extension to question answering.
Therefore, in this paper, 
\begin{itemize}
\item We closely analyze the differences between VQA and image-based CQA tasks, and identify the challenges in multimodal CQA that may hinder the performance of VQA models.
\item Following this, we propose modifications to VQA-inspired models for a better CQA-task performance. Our key contributions include learning an additional global weight for image in the image-text combination step and introducing auxiliary tasks to learn better grounding features. 
\item We evaluate our model against baselines from text-only \& VQA models, and other frequently used methods for image-text combination, on the Chiebukuro dataset.
\item Finally, we use an ablation study to quantify the contributions of each of our suggested changes. We will also be making our source code for the models publicly available. 
\end{itemize}    

A natural counterpart to VQA is answering image-based CQA questions. However, this is far more difficult and subjective compared to answering in VQA due to varying answer lengths and composition, requirement of non-trivial external knowledge that must be modified according to the question's context, and necessity of human opinions. Therefore, we do not tackle answering in this work, but remain optimistic about our ability to use the results, inferences, and models from this work to answer a subset of simpler factoid-based CQA image questions in our future work. 

To the best of our knowledge, our work is the first to tackle the challenges of multimodal CQA, and also to adapt VQA models for tasks on a more ecologically valid source of visual questions posted by humans seeking the expertise of the community (as opposed to straightforward questions that query the image). It is worth noting that \cite{tamaki2018classifying} deals with the same dataset source, comparing joint embedding methods for a basic classification task, but does not attempt to address any CQA-specific challenges. By identifying and targeting such specific idiosyncrasies, we get a $>8$\% jump in classification accuracy, and $>25\%$ relative MRR increase on expert retrieval compared to the model in \cite{tamaki2018classifying}.
% \todo{change citations containing et al.}

% \fujita{You can not omit the next paragraph as this is a sort of required style for a long paper. Instead, you may omit some sentences from above two paragraphs.}
% The rest of the paper is structured as follows: we start with a brief overview of work in related domains in Section \ref{sec:related_work}. In Section \ref{sec:vqa_cqa_diff}, we explore the differences between VQA and CQA, understand the CQA tasks, and identify the potential failure points of applying VQA methods directly to them. In Section \ref{sec:our_sol}, we detail our proposed solutions, and provide our final model's description in Section \ref{sec:final_model}. Finally, we present experiments comparing our model against the baselines, along with an ablation study and qualitative examples to better understand our contributions.

\begin{table*}[t]
  \centering
  \captionsetup{justification=centering}
  \caption{\textbf{Statistics for VQA (English) and YC-CQA (Japanese) datasets.} }
  \renewcommand{\arraystretch}{1.15}
  \begin{tabular}{|>{\centering\arraybackslash}m{5em}|>{\centering\arraybackslash}m{4.5em}|>{\centering\arraybackslash}m{5em}|>{\centering\arraybackslash}m{4.5em}|>{\centering\arraybackslash}m{6.25em}|>{\centering\arraybackslash}m{6.25em}|>{\centering\arraybackslash}m{4.5em}|>{\centering\arraybackslash}m{5.25em}|}
  \hline
    Dataset & Total \# of Questions & Question Text's Vocab Size & Answer Text's Vocab Size & Avg. Question Length (\#words)& Avg. Answer Length (\#words) &  Total \# of Question Categories & Average Categories per Question\\
    \hline 
    DAQUAR  & 12,468 & 2520 & 823 & 11.53 & 1.15 & 3 & 1.00\\
    \hline
    COCO-QA & 117,684 & 12,047 & 430 & 8.65 & 1.00 & 4 & 1.00 \\ 
    \hline
    VQA v2 & 658,111 & 26,749 & 29,548 & 6.20 & 1.16 & 65 & 1.00\\
    \hline
    YC-CQA & 1,018,833 & 176,921 & 335,658 & 71.54 & 62.15 & 38 & 2.73 \\ 
    \hline
  \end{tabular}
  
  \label{tab:1}
\end{table*}

%%%%%%%%%%%%%%%%%%%%%%%%%%%%%%%%%%%%%%%%%%%%%%%%
\section{Related work} \label{sec:related_work}

% \paragraph{Automatic Question Classification for CQA service}
%The CQA task of automatic question classification is useful for tagging newly posted questions and suggesting an appropriate question category to the asking user.
% and also acts as a good first step towards potentially answering the question. 
%This has led to many studies focusing on this kind of classification \cite{zhang2003question}, \cite{saha2013discriminative}, \cite{stanley2013predicting}, \cite{tamaki2018classifying}.
% \fujita{Please briefly explain each of \cite{zhang2003question}, \cite{saha2013discriminative}, \cite{stanley2013predicting}, \cite{tamaki2018classifying}.}

% \paragraph{Expert Retrieval for CQA service}
% Another useful problem to solve is that of retrieving ``experts". Here, the aim is to identify and retrieve users from the community who are likely to provide answers to a given question. This provides an efficient way to make the community well-knit, provide better content to askers, and recommend only the relevant questions from a gigantic pool of queries to the potential answerers. Previous works include \cite{zheng2017deep}, \cite{zhao2016expert}, \cite{riahi2012finding}, \cite{liu2005finding} among others.
% \fujita{Please briefly explain each of \cite{zheng2017deep}, \cite{zhao2016expert}, \cite{riahi2012finding}, \cite{liu2005finding}.}

\textbf{CQA Tasks}. Initial approaches for question categorization and expert retrieval were heavily based on supervised machine learning approaches utilizing hand-crafted features (\cite{stanley2013predicting}, 
% \cite{zhang2003question},
\cite{saha2013discriminative}, \cite{roberts2014automatically}) language models (\cite{balog2006formal}, \cite{riahi2012finding}, \cite{zheng2012algorithm}), topic models (\cite{tian2013predicting}, \cite{zhou2009routing}, \cite{yang2013cqarank}), and network structure information (\cite{zhao2016expert}, \cite{srivastava2017soft}, \cite{singh2011cqc}). A recent shift to end-to-end deep learning approaches (\cite{zheng2017deep}, \cite{tamaki2018classifying}, \cite{kim2014convolutional}, \cite{severyn2015learning}) has shown successful results for these tasks and the related task of question similarity ranking. Apart from \cite{tamaki2018classifying}, all works are focused on only the text modality. 

\textbf{Joint Image-Text Representation Learning}.  Most prevalent application of works combining  computer vision (CV), natural language processing (NLP), and knowledge representation \& reasoning (KR) have been in image captioning and VQA tasks. While many methods for image captioning use the image representation as context for the decoder segment (\cite{you2016image}, \cite{chen2015mind}), it can also be casted into a encoder-decoder framework that uses joint image-text representations \cite{kiros2014unifying}. We choose to instead focus on VQA for a number of reasons. First, there has been significantly more progress in VQA due to hardness of evaluation of the image captioning task, along with the the fact that captioning task lacks the need for reasoning and requires only a single coarse glance at the image \cite{antol2015vqa}. Second, the usage of joint representations in VQA (classification over answers) is more similar to our use case of classification and ranking, compared to text generation in captioning. Third, many recent works have developed models that can be used for both captioning and VQA (\cite{wu2018image}, \cite{anderson2018bottom}). Therefore, there seems no apparent reason to favor captioning, and we streamline our work by focusing on VQA methods.

% Image Captioning with Semantic Attention
% Mind’s Eye: A Recurrent Visual Representation for Image Caption Generation
% Unifying Visual-Semantic Embeddings with Multimodal Neural Language Models
% Image Captioning and Visual Question Answering Based on Attributes and External Knowledge
% Bottom-Up and Top-Down Attention for Image Captioning

% Question categorization and expert retrieval have been extensively research topics in the field of community question answering owing to their contribution in improving the accessibility and effectiveness of the platform for the users.

% \paragraph{Combining Text and Images in VQA}
\textbf{VQA Methods}. The two most common approaches for combining image-text representations in VQA are joint embedding methods and attention-based mechanisms \cite{wu2017visual}. 
% The former aims to derive representations in a common space that allows performing inference over text and image contents. 
One of the most basic methods is to simply concatenate the derived text and image embeddings \cite{zhou2015simple}. A better method is concatenation of the element wise sum and product \cite{saito2017dualnet}. A bilinear pooling-based method was found to be effective by \cite{fukui2016multimodal}. 
% With respect to the VQA task, these approaches suffer from the shortcoming of taking in global image-wide features, which may obscure task-relevant regions and feed irrelevant and noisy information at the prediction stage. 
Attention based approaches learn a convex combination of spatial image vectors as the contributor to the final joint embedding. A simple and popular attention model for VQA is the stacked attention network \cite{yang2016stacked}, where the text is seen as a query for retrieving attention weights for image regions. Methods involving attention over both image and text include \cite{lu2016hierarchical}, \cite{nguyen2018improved} among others.
% The element wise sum of the weighted image vector and the text vector can either be used as the final joint embedding, or can be used as a refined query to learn additional layers of attention on the image for multi-step reasoning.
% The hierarchical image-text co-attention model introduced by \cite{lu2016hierarchical} learns attention over both image and text regions at the word, phrase and sentence levels. 
\cite{fukui2016multimodal} also describes a version of its model that utilizes attention. More recent improvements in performance come from use of bottom up attention features \cite{anderson2018bottom}, intricate attention mechanisms like bilinear attention maps \cite{kim2018bilinear}, and careful network tuning and data augmentation methods \cite{jiang2018pythia}.
% Attention mechanisms are discussed in more detail in the section on addressing CQA challenges.

% Simple Baseline for Visual Question Answering
% Improved Fusion of Visual and Language Representations by Dense Symmetric Co-Attention for Visual Question Answering

\section{Understanding VQA-CQA Differences} \label{sec:vqa_cqa_diff}
It is crucial to understand the differences between the question-image pairs in VQA and CQA in order to identify the unique challenges posed by the new dataset and address them by means of appropriate modifications. The dataset consists of questions posted over an year on YC-CQA
\footnote{Data available for purchase from \url{https://nazuki-oto.com/chiebukuro/index.html}},
which allows questions with and without an image. In this work, we only deal with questions accompanied by an image. Table \ref{tab:1} presents a simple comparison highlighting different aspects of the data, contrasting YC-CQA with the most commonly used VQA datasets. To better understand the contrast, we first analyze the quantitative differences and next discuss some of the more fundamental differences that are the driving influences for our proposed modifications.

\subsection{Quantitative Dataset Differences}
\begin{figure}[t]
\centering
\begin{subfigure}[b]{.24\textwidth}
  \centering
  \includegraphics[width=0.97\linewidth]{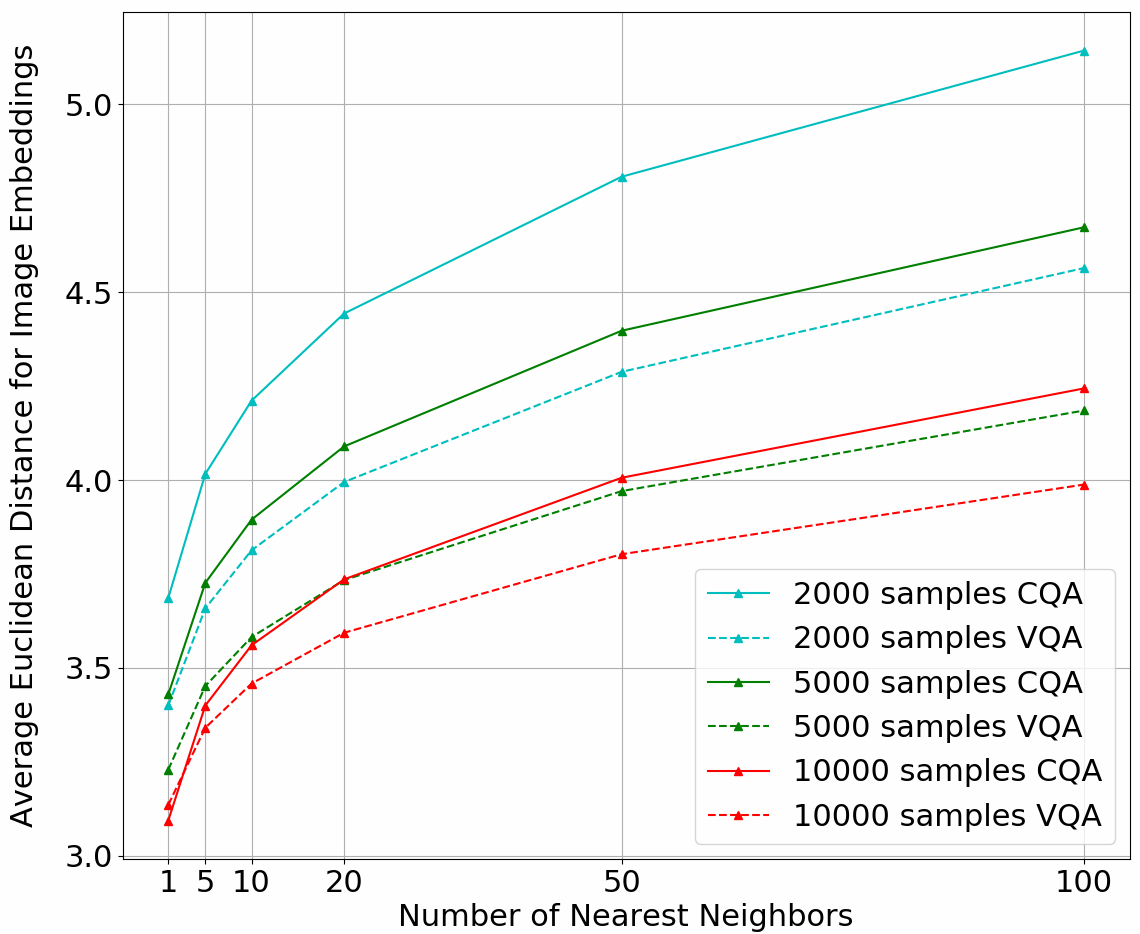}
  \label{fig:2a}
\end{subfigure}%
\begin{subfigure}[b]{.24\textwidth}
  \centering
  \includegraphics[width=0.97\linewidth]{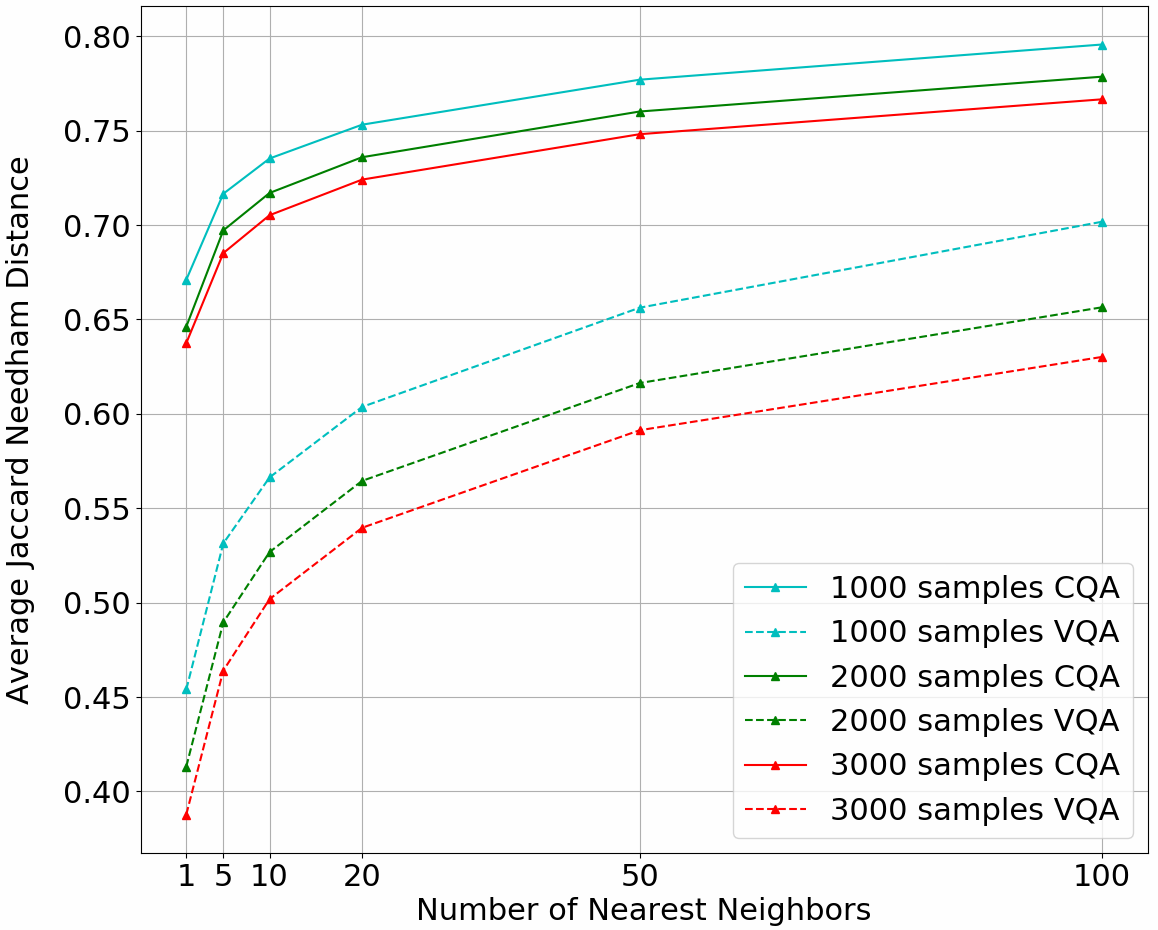}
  \label{fig:2b}
\end{subfigure}
\caption{Image \& text sample proximity in CQA vs VQA. (a): Mean average distance to K-nearest neighbors for image representations; (b): Mean average distance to K-nearest neighbors for BoW text representations.}
\label{fig:2}

\end{figure}

Table \ref{tab:1} highlights the complexity of CQA data in terms of a significantly larger vocabulary set and average question and answer lengths (despite different languages, the magnitude of difference is sufficient to drive the point). The CQA dataset presents a significantly higher noise in its text and image data. Common methods for question generation in VQA are to either automatically convert image caption data into questions or to have human annotators produce the questions on the basis of predefined guidelines. 
These lead to a sense of homogeneity in the questions -  one that is missing in CQA, where question authors comprise a large number of different individuals. 
These differences are partly quantified by our experiment, where we select a subset of samples and retrieve the nearest neighbors for each text sample using Jaccard-Needham dissimilarity as the distance metric. We compare the mean average distance of the neighbors for the VQA and YC-CQA datasets. The result of this experiment, performed for randomly sampled 1k, 2k, and 3k sized subsets, demonstrates closer distances between similar set of questions in VQA than in CQA (Figure \ref{fig:2}(b)).

The image diversity in CQA is expected to be greater as well. Images in the DAQUAR \cite{malinowski2014nips} dataset comprise indoor scenes, COCO-QA \cite{ren2015exploring} images contain common-objects-in-context, and the VQA dataset \cite{antol2015vqa} contains abstract scenes and clipart images. All these categories are subsumed by the images on the CQA platform, as users are not restricted in terms of the type or attributes of the image and the question they post. The results for an experiment similar to the one for texts, but using pre-trained ResNet-derived embeddings for sampled images, is shown in Figure \ref{fig:2}(a).

\subsection{CQA Tasks}
To clarify further differences, we first take a closer look at our two intended end-tasks.
\subsubsection{Category Classification}
The category assignment for a question is provided by the asking user. The available category choices are arranged in a hierarchical fashion, with each category having a single parent category. The number of level-0 categories is 14, followed by 95 level-1 and 415 level-2 categories. Each question's most specific category can come from any of these levels, with the condition that a level-1 or level-2 category labeled question is also labeled with the parent category. For example, for the category hierarchy `\textit{Life Sciences $>$ Plants \& Animals $>$ Plants}', the possible category assignments are `\textit{Life Sciences}', or `\textit{Life Sciences}' \& `\textit{Plants} \& \textit{Animals}', or `\textit{Life Sciences}' \& `\textit{Plants \& Animals}' \& `\textit{Plants}'.

Most level-2 and many level-1 categories have an extremely sparse presence in the dataset. Little training data is available, and for practical reasons it makes sense to skip such rare categories and settle for predicting only their parent category. Our final classification is done on 38 categories, selected by eliminating ones occurring in less than 5k samples. We treat this as a flat multi-label classification problem. Thus, a question tagged as `\textit{Life Sciences} $>$ \textit{Plants \& Animals}' is labeled as both `\textit{Life Sciences}' and `\textit{Plants \& Animals}'. This also leads to lower training loss for over- and under-generalized predictions compared to completely wrong ones.

\subsubsection{Retrieving Experts} 
% Obtaining fast and relevant answers to questions is dependent on recommending the right questions to frequent answerers or ``experts". 
We define our candidate pool of experts to contain users with more than 50 answers in the initial six-month period from which our dataset is drawn. Therefore, the relevant set of experts for a question is comprised of users that both answered the question and are present in the candidate pool. Similar to  \cite{liu2005finding}, we use mean reciprocal rank (MRR) as the evaluation measure. Note that in practical settings, we append features such as `last active time of user', `asking user's reputation' etc. to our feature set , but here we compare the models on their ability to retrieve experts based solely on text-image pairs of questions answered by users. This makes sense since the information contribution from other features is mostly orthogonal to this.

\subsection{Identifying Fundamental Challenges}\label{subsec:challenges}
While more noise in the data poses a problem to any learning model, it is important to identify more pressing CQA challenges that question the fundamental assumptions of VQA models. 

The first challenge stems from the difference in the role of images . In VQA, the question generally queries the image, and it is imperative to gain a visual understanding in order to answer it. This strong dependence on image is almost completely non-existent in CQA tasks for most questions. For many samples the text contains enough information to successfully perform the task, and/or the image contains relatively little information, and at times is just posted as a placeholder, or is irrelevant for gaining an understanding of the question. 
% This can be seen in our results, where joint embedding approaches do not give a major improvement (seen relative to added image data available) over the text-only baseline (${\sim}4\%$ classification accuracy increase).
The combination of image and text embeddings in VQA models has the implicit assumption of balanced information content from both for the end task. Therefore, to deal with information imbalance between text and image in CQA samples, our first intended modification is to model this difference by learning an additional global weight for the image, which would signify its contribution towards the final joint embedding. 

%\begin{figure}[t]
% \centering
% \begin{subfigure}[b]{.35\textwidth}
%   \centering
%   \includegraphics[width=0.97\linewidth]{KNN_image}
%   \label{fig:2a}
%   \caption{Mean average distance to K-nearest neighbors for image representations}
% \end{subfigure}\\\vspace{0.3cm}
% \begin{subfigure}[b]{.35\textwidth}
%   \centering
%   \includegraphics[width=0.97\linewidth]{KNN_text}
%   \label{fig:2b}
%   \caption{Mean average distance to K-nearest neighbors for BoW text representations}
% \end{subfigure}
% \caption{Image \& text sample proximities in CQA vs VQA.}
% \label{fig:2}

% \end{figure}

The second challenge is to correctly ground the text to image relation. Grounding here implies pairing the relevant objects or regions in an image to the corresponding references to them in the accompanying text. VQA questions are mostly single sentences with keywords referring to objects in the image, and the final answer is dependent on such references. This leads to sound learning of grounding features. This is much more difficult for CQA because i) large question texts hamper identification of text regions where the image is referred, ii) low contribution of image towards the final task means that the model tends to skip grounding, and iii) the CQA tasks are simpler compared to VQA, as VQA needs the multimodal features to interact, while CQA tasks tend to focus more on textual features, thus impeding the model's ability to learn grounding features well. 
Thus, our second intended modification is to design tasks that help to learn these features better, and to use those features to improve performance on our main tasks. 
% Considering the large amount of contextual information and redundancy in CQA question texts, a broad scenario where we expect to benefit from the image data is when multiple topics can be inferred from the textual features. In such cases,  identifying the terms in the text that refer to the image can help the model to understand the question's subject better, and improve the image-text combination embedding.
Intuitively, with this modification, we stand to benefit in the scenarios where multiple topics can be inferred from the textual features. In such cases,  identifying the terms in the text that refer to the image can help the model to understand the question's subject better, and improve the image-text combination embedding.

Another intuitive observation is that attention cannot be expected to give the same impressive improvements for CQA tasks as it does for VQA. The reasons are similar: attention is suited for VQA, where salient characteristics of different image regions play differently important roles in both understanding the question and inferring the answer. For CQA, along with absence of such dependencies, poor grounding makes it harder to learn good attention weights. Therefore, solving the second challenge can also help to utilize the attention mechanism a bit better.

\section{Addressing CQA Challenges} \label{sec:our_sol}

% Different VQA models typically differ in the approach they take in modeling one or more of these three components: i) deriving high-level image representation, ii) deriving high-level text representation, and iii) a combination method for these two to derive the final representation. This representation is fed to a classification layer to derive the answer. We can easily replace the final layer to perform final tasks of our choice as long as they are dependent on the information in the text and the image. Therefore, we focus on the two identified challenges - image's contribution towards the task, and difficulty in learning grounding features - and present our proposed solutions as follows:
Now we present our solutions for the two identified challenges - varying information contribution from images across samples, and difficulty in learning grounding features.

\subsection{Learning a Global Image Feature Weight}

The strong image dependence in VQA makes it feasible to use methods such as element-wise sum-product and concatenation for image and text representations. Attention mechanisms learn attention weights for different image regions that are derived using a final softmax layer and so, sum to 1. These methods, however, provide no way for the model to learn to weigh the contributions of text and image separately for each sample, which becomes important for CQA, where these two contribute significantly different amounts of information in different samples. We therefore introduce the learning of a global weight for image, both with and without attention.

\subsubsection{Global Weight w/o Attention} \label{sec:global_image_wo_attention} Given the text and image vectors, we want to learn a parameter $\alpha$ that acts as the scalar weight for the image vector's contribution. This parameter is derived by contribution of both the derived image vector $\mathbf{v_I} \in {\rm I\!R}^{d}$ and the derived text vector $\mathbf{v_T} \in {\rm I\!R}^{d}$ , as: 
\begin{align}
\mathbf{h}_A &= \tanh(\mathbf{W}_{IA}\mathbf{v}_I + \mathbf{W}_{TA}\mathbf{v}_T + \mathbf{b}_A), \\
\alpha &= \sigma(\mathbf{W}_{A\alpha}\mathbf{h}_A + b_{A\alpha}),
\end{align}
where $\mathbf{W}_{IA}$, $\mathbf{W}_{IA} \in {\rm I\!R}^{k \times d}$, and $\mathbf{W}_{A\alpha} \in {\rm I\!R}^{1 \times k}$. 
Simply multiplying $\alpha$ with $\mathbf{v}_I$ to get the image contribution $\mathbf{\widetilde{v}}_I$ can be problematic for the image-text embedding product we plan to use in the joint embedding in Eq. \ref{eq:8}. Therefore, we distribute the $\alpha$ and $1 - \alpha$ parameters between $\mathbf{v}_I$ and a `fall back' option $\mathbf{v}_{T'}$ obtained by a non-linear transformation on $\mathbf{v}_T$: 
\begin{equation}
\mathbf{v}_{T'} = \tanh(\mathbf{W}_{TT'}\mathbf{v}_T + \mathbf{b}_{T'}),
\end{equation}
\begin{equation}
\mathbf{\widetilde{v}}_I = \alpha*\mathbf{v}_I + (1 - \alpha)*\mathbf{v}_{T'}
\end{equation}

The final image-text embedding is derived as
\begin{align} 
\mathbf{v}_{IT} &= [\mathbf{v}_T + \mathbf{\widetilde{v}}_I, \mathbf{v}_T * \mathbf{\widetilde{v}}_I] \label{eq:8}
\end{align}

\subsubsection{Global Weight with Attention} \label{subsec:globalatt}
Given the spatial image embedding $\mathbf{v}_{spI} \in {\rm I\!R}^{d \times m}$, where \textit{d} is the representation dimension for \textit{m} image regions,  attention weights and image contribution in \cite{yang2016stacked} are derived as:
\begin{align}
\mathbf{h}_A &= \tanh(\mathbf{W}_{IA}\mathbf{v}_{spI} \oplus (\mathbf{W}_{TA}\mathbf{v}_T + \mathbf{b}_A )),\\
\bm{\alpha}_I &= \text{softmax}(\mathbf{W}_{A\alpha} \mathbf{h}_A + \mathbf{b}_\alpha ),\\
\mathbf{\widetilde{v}}_I &= \Sigma_i \alpha_i\mathbf{v}_{spI_i}, \label{eq:11} 
\end{align}
where $\mathbf{W}_{IA}$, $\mathbf{W}_{TA} \in {\rm I\!R}^{k \times d}$, 
$\mathbf{W}_{A\alpha} \in {\rm I\!R}^{1 \times k}$,  $\oplus$ denotes the addition
of a matrix and a vector. 

To introduce the global image weight, we adopt an approach similar to the one used in \cite{lu2017knowing}.
% for image captioning,  where one of the attention weights decides the image representation's input towards the output at each time step of the decoder RNN. 
% Therefore, for the
For the \textit{m} image regions, instead of learning 
\{$\alpha_1, \alpha_2, \ldots, \alpha_m$\} attention weights with $\sum_{i=1}^m \alpha_i = 1$, we learn an additional weight $\alpha_{m+1}$ such that now $\sum_{i=1}^{m+1} \alpha_i = 1$. This allows the model to attribute more weight to $\alpha_{m+1}$ (assigned to $\mathbf{v}_{T'}$) when the image contribution is determined to be low. The attention weights are derived as follows:
\begin{align}
\mathbf{v'}_{spI} &= [\mathbf{v}_{spI}, \mathbf{v}_{T'}],\\
\mathbf{h'}_A &= \tanh(\mathbf{W}_{IA}\mathbf{v'}_{spI} \oplus (\mathbf{W}_{TA}\mathbf{v}_T + \mathbf{b}_A )),\\
\bm{\alpha'}_I &= \text{softmax}(\mathbf{W}_{A\alpha} \mathbf{h'}_A + \mathbf{b'}_\alpha ),
\end{align}
where $\mathbf{v'}_{spI} \in {\rm I\!R}^{d \times (m+1)}$, $\mathbf{h'}_{A} \in {\rm I\!R}^{k \times (m+1)}$, $\bm{\alpha'}_{I} \in {\rm I\!R}^{1 \times (m+1)}$. 
The image contribution $\mathbf{\widetilde{v}}_I$ is derived as in equation \ref{eq:11} with $\mathbf{v'}_{spI}$ replacing $\mathbf{v}_{spI}$. The joint embedding is then obtained as in equation \ref{eq:8}.

\begin{figure*}[ht]
\begin{subfigure}[b]{0.47\textwidth}
\centering
\includegraphics[width=\linewidth,height=3.5cm]{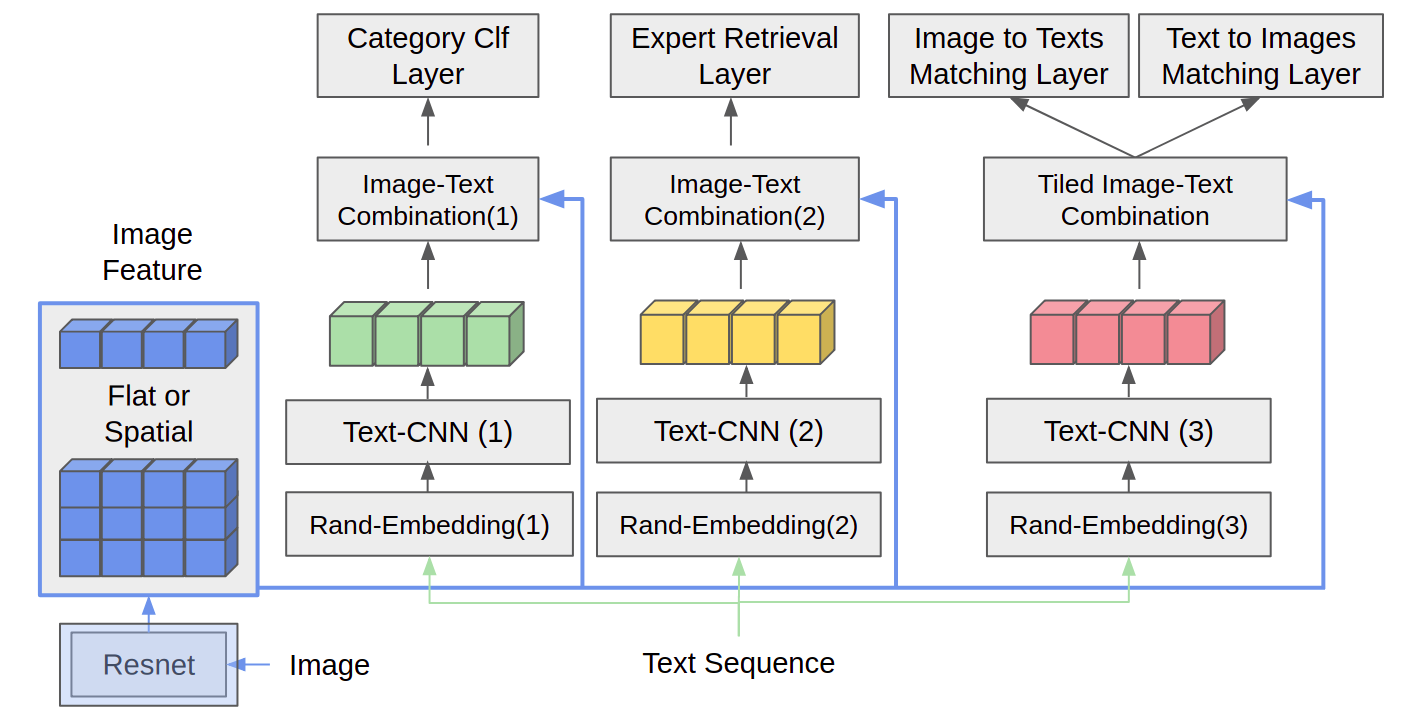}
\caption{Different weights are learned for deriving the text embedding and for the image-text combination layer for each task (classification, retrieval, and auxiliary (sec \ref{sec:aux})). The image feature input is a flat embedding for models w/o attention, and spatial for attention-based ones.}
\label{fig:trainsub1}
\end{subfigure}
\begin{subfigure}[b]{.435\textwidth}
\centering
\includegraphics[width=\linewidth,height=3.75cm]{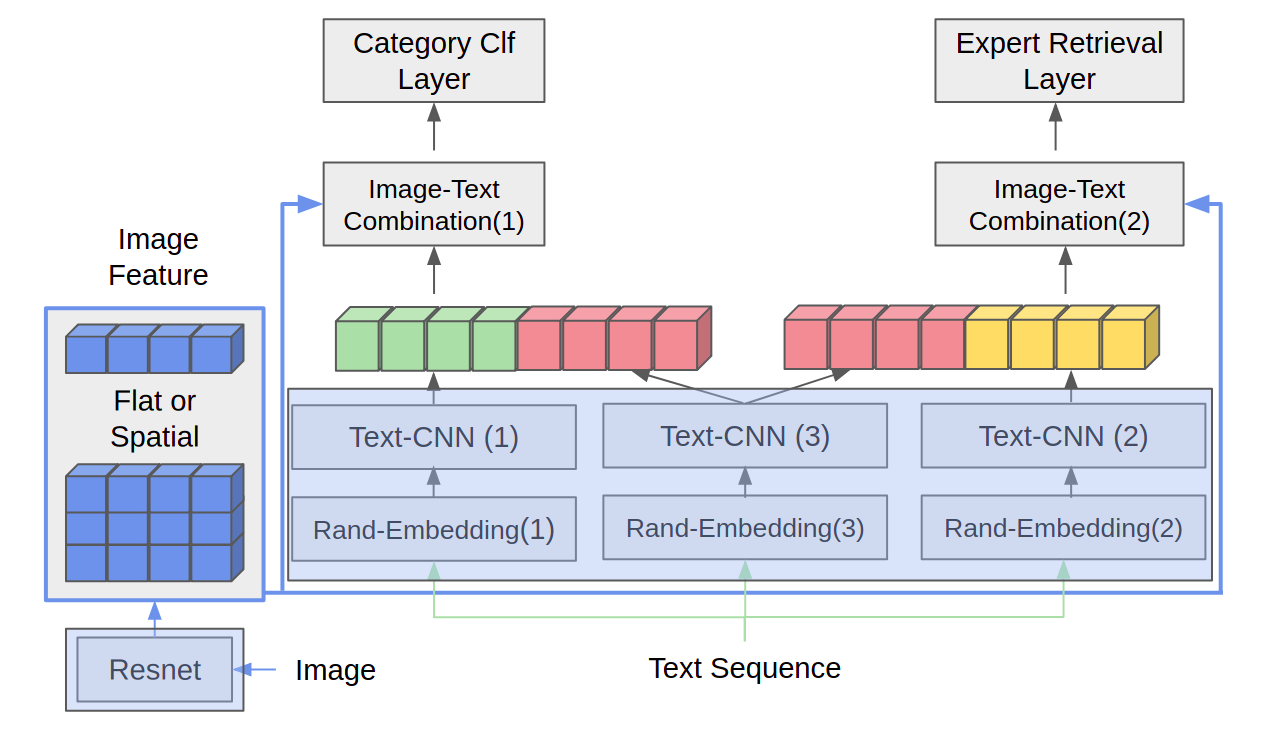}
\caption{The word embeddings and the text CNN filters for each model are fixed. The text embedding from auxiliary task's text CNN (red) is concatenated with its corresponding text embedding in the main tasks (green and yellow). Only the parameters in the image-text combination layers are learned in this step.}
\label{fig:trainsub2}
\end{subfigure} \\\vspace{0.2cm}
\begin{subfigure}[b]{.465\textwidth}
\centering
\includegraphics[width=\linewidth,height=3.7cm]{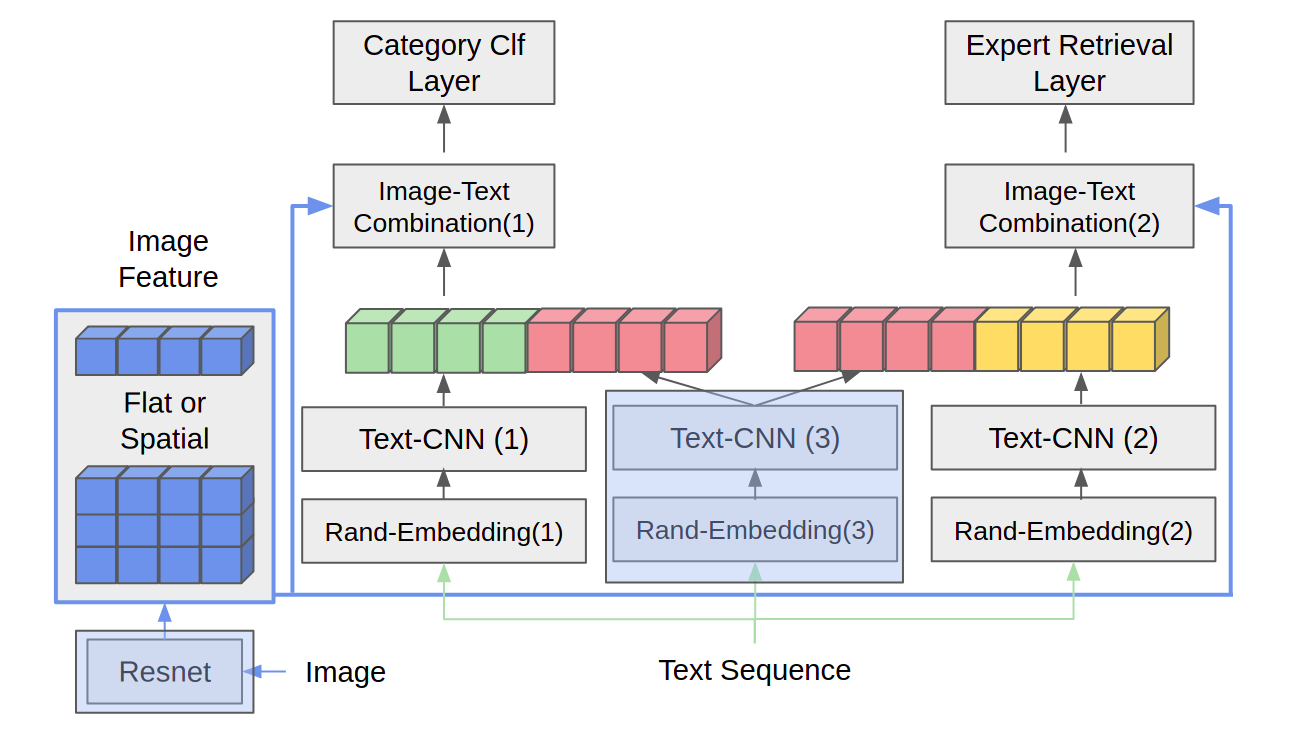}
\caption{Finally, fine-tuning is allowed in the weights for text embedding derivation in the two main tasks. This helps to update filter weights to better identify words that can be referring to salient characteristics of the associated image.}
\label{fig:trainsub3}
\end{subfigure}
\caption{Training pipeline for auxiliary and final tasks. Presence of the translucent light blue layer implies ``frozen" weights, i.e. absence of backpropagation and weight updates through those channels.}
\label{fig:4}

\end{figure*}

\subsection{Learning Grounding Features through Auxiliary Tasks} \label{sec:aux}
We discussed the problem of failing to learn grounding features in CQA. Using hints, i.e., predicting the features as an auxiliary task, is one of the proposed approaches for the problem of learning features that might not be easy to learn using only the original task (\cite{abu1990learning}, \cite{ruder2017overview}), with success shown in recent work on sentiment analysis in \cite{yu2016learning} and on name error detection in \cite{cheng2015open}. We propose two auxiliary tasks to learn better grounding features and outline the training pipeline for utilizing these towards the final tasks.

\subsubsection*{Image-Text Matching Auxiliary Tasks}
A comparatively more challenging task on the CQA data is matching a question's image to its corresponding text from among a pool of candidate texts, and vice-versa. To do this well, it's necessary to learn the regions in the text that refer to salient regions in the image - providing an effective logical solution to the problem of poor grounding. Furthermore, this task relies simply on clever data usage for training, requiring no extra labels or samples. 
% We try to encode this information in the text embeddings using explicit auxiliary tasks. 

Formally, given our image-text questions dataset $\mathcal{D} = \{(\text{Im}_i, T_i)\}_{i=1}^{N}$, where $\text{Im}_{i}$ and $T_i$ are the associated image and text with the $i^{th}$ question, respectively, we construct two new training sets for the image-to-texts and text-to-images matching tasks. For the former, we set up the task as follows: given a question image and five candidate texts, the aim is to correctly identify the question text corresponding to the image among the candidates. We construct $ \mathcal{D}^{IT} = \{\text{Im}_i, \mathbf{T}_{\text{cand}_i}\}_{i=1}^{N}$, where $\mathbf{T}_{\text{cand}_i} = \{T_{i1}, T_{i2}, T_{i3}, T_{i4}, T_{i5}\}$ such that $T_i\in\mathbf{T}_{\text{cand}_i}$, and the other four texts are negatively sampled. Similarly, for text-to-images matching, we construct $\mathcal{D}^{TI} = \{\textbf{Im}_{\text{cand}_i}, T_i\}_{i=1}^{N}$, s.t. $\textbf{Im}_{\text{cand}_i} = \{\text{Im}_{i1}, \text{Im}_{i2}, \text{Im}_{i3}, \text{Im}_{i4}, \text{Im}_{i5}\}$ and $ \text{Im}_i \in \textbf{Im}_{\text{cand}_i}$. The training is described in subsections \ref{sub:final_layers} and \ref{sub:training_pipeline}.

\section{Final Model Description} \label{sec:final_model}

We now present the full picture of our model which utilizes the solutions we have proposed.

\subsection{Text Representation}

The text data is in Japanese language. We do some elementary preprocessing by removing HTML characters and replacing URLs with a special token. Tokenizing Japanese text is challenging 
since 
words in a sentence aren't separated by spaces. Therefore, we use the morphological analyser Janome\footnote{http://mocobeta.github.io/janome/en/} for word splitting.  

We use randomly initialized word embeddings (trained end-to-end, similar to \cite{yang2016stacked}), followed by a CNN-based architecture from \cite{kim2014convolutional} to derive the high-level text representation $\mathbf{v}_T$. CNN-based architectures have shown successful results in previous VQA works (\cite{yang2016stacked}, \cite{ma2016learning}, \cite{lu2016hierarchical}), and can be particularly useful for extracting features important for CQA tasks. We learn filters of sizes 1, 2, and 3 over the sequence of embeddings 
with max-pooling over each full-stride of a filter to obtain the text representation $\mathbf{v}_T$.

\subsection{Image Representation}

Most VQA works use networks pre-trained on ImageNet such as the ResNet \cite{he2016deep} or VGGNet \cite{simonyan2014very}. Here, we use the pre-trained ResNet network, utilizing the final spatial representation for attention-based networks, and the final flat embedding for other models. Images are resized to 224 x 224, giving a 7 x 7 x 2048 dimensional spatial embedding $\mathbf{v}_{spI}$, and 2048-D flat embedding $\mathbf{v}_I$.

\subsection{Joint Representation and Final Layers} \label{sub:final_layers}

We use the global image weight with attention mechanism from Section \ref{subsec:globalatt} to get the joint embedding $v_{IT}$. $v_{IT}$ can be input to different final layers for different tasks, which are described below.

\subsubsection{Category Classification Layer}
A fully-connected layer with sigmoid activation is used for the multi-label classification task. 

\subsubsection{Expert Retrieval Layer} 
\begin{figure}[t]
  \centering
    \includegraphics[width=0.82\linewidth, height=3.6cm]{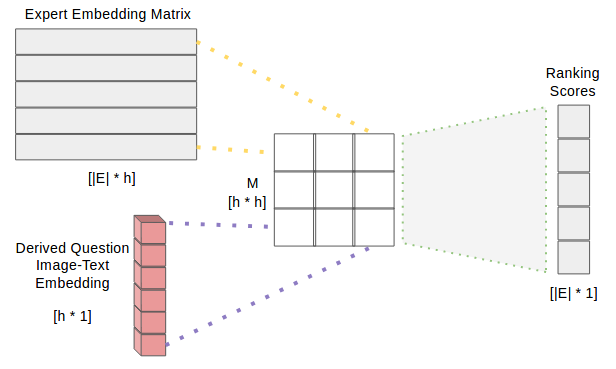}
    \caption{Final layer for ranking experts, given $|E|$ candidates and embedding size $h$. The question's image-text embedding (red) is derived using any of the VQA-based/adapted model, and then multiplied with the similarity matrix and expert embedding matrix to get a score for each candidate expert.}
    \label{fig:arch1}
\end{figure}

For expert retrieval, we try to score each candidate expert for each given question. Hence, we use an architecture inspired by \cite{severyn2015learning}, using a matching matrix to score the candidate pool. Formally, given the joint embedding $\mathbf{v}_{IT} \in {\rm I\!R}^{h}$, for an expert with embedding representation $\mathbf{e}_i \in {\rm I\!R}^{h}$ (randomly initialized, learned end-to-end), the score for this expert is:
\begin{align}
\text{match}(\mathbf{e}_i, \mathbf{v}_{IT}) = \mathbf{e}_i^T\mathbf{M}\mathbf{v}_{IT},
\end{align}
where $\mathbf{M} \in {\rm I\!R}^{h\times h}$ is the randomly initialized, end-to-end learned matching matrix as shown in Figure \ref{fig:arch1}.
% \fujita{Explain Figure 3}

\subsubsection{Auxiliary Tasks} 
\begin{figure}[t]
  \centering
    \includegraphics[width=0.78\linewidth, height=3.2cm]{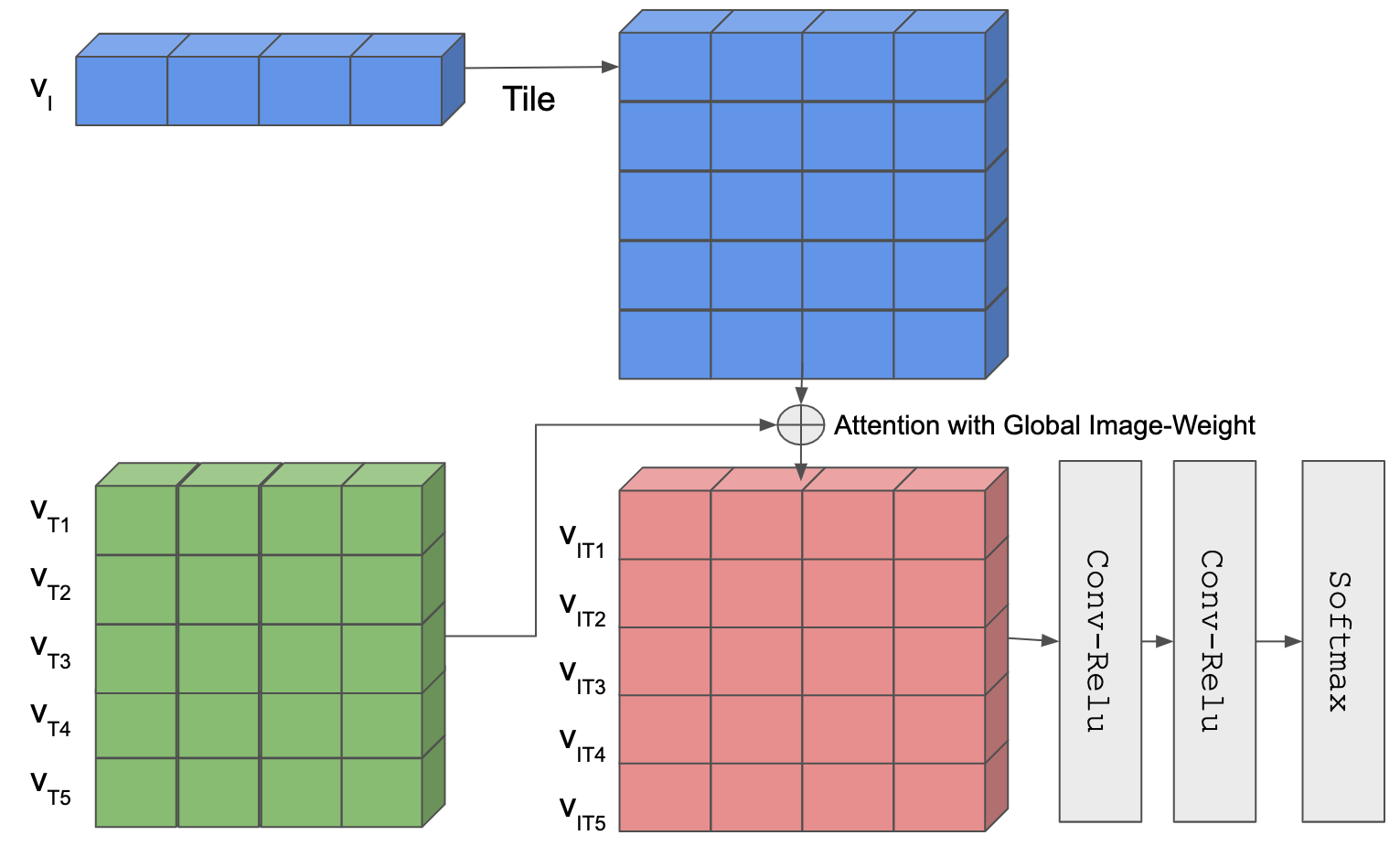}
    \caption{Architecture for image-to-texts matching auxiliary task. The image representation (blue) is tiled and combined with five candidate text representations (green) to derive matching scores for the image with each of the five texts. }
    \label{fig:arch2}
\end{figure}

This is a five-class single-label classification task. Five joint embeddings are derived since each sample has either five candidate images or five candidate texts. The prediction of the correct candidate uses the architecture shown in Figure \ref{fig:arch2} for image-to-texts matching task. The combined representations (red in the Figure) are passed through two convolutional layers to derive five scores at the end. Softmax is applied to these score to obtain probabilities over the candidates. For the text-to-images matching task, the roles of image and text in Figure \ref{fig:arch2} are simply reversed.
% \fujita{Explain Figure 4}

\subsection{Training Pipeline}\label{sub:training_pipeline}
Figure \ref{fig:4} shows the training pipeline. The three steps are:
\begin{enumerate}
\setlength\itemsep{0.1em}
\item First, the two main tasks and the auxiliary tasks are individually trained. For the auxiliary tasks, depending on the task being optimized for the current batch, either the text or image input is five-fold the batch size. Since the other input is tiled to be quintupled, the rest of the architecture (apart from input) remains the same for the two tasks, 
% (Fig \ref{fig:arch2} for image-to-texts matching auxiliary task, for the text-to-images matching auxiliary task, the roles of the image and text embeddings in Figure \ref{fig:arch2} are simply reversed.),
and the two losses are optimized without any scaling (Figure \ref{fig:4} (a)).
\item `Freezing' the text embeddings and text CNN for all three, and training the classification and retrieval models using text embeddings derived from concatenation of the original and the ones from auxiliary tasks' text CNN (Figure \ref{fig:4}(b)). The parameter sizes of the model can be changed to take in double the usual text embedding size, or an FC layer can be used to reduce the dimensions to half. Both approaches produced similar results in our experiments.
\item After a sufficient number of epochs (25 in our experiments), we fine-tune the text CNN for the main tasks to gain further minor improvements, as shown in Figure \ref{fig:4}(c).
\end{enumerate}

\section{Experiments}

\subsection{Setup}
% We use TensorFlow \cite{tensorflow2015-whitepaper} to develop our models.
For both tasks, we use 80\%-10\%-10\% splits for training, validation and test sets respectively.
Batch size is 128 for main tasks and 32 for auxiliary tasks (due to the five-fold inputs). Training components include early stopping, learning rate \& weight decay, and gradient clipping. For images, basic data augmentation and flipping is applied. For the text model, embedding size of 128 and filter sizes of 128 for 1-gram, and 256 for 2- and 3-grams worked best.  More detailed notes for exact reproducibility will be outlined in our code release.

% The final spatial representation used the 7x7x2048-D output from the last pooling layer of the pre-trained ResNet, and the flat image embedding was derived by averaging the spatial representation across the regions to get a 2048-D output. For baseline models, transforming both image and text dimensions to 1024-D (using FC layer for image and text flat embeddings, and 1x1 convolutions for spatial image feature matrix) worked well. 

% The final results were obtained after finding the best intermediate dimensions through a small grid search.

\subsection{Results and Analysis}
For both tasks, we use the following models as baselines:
\begin{itemize}
\item \textbf{Random}: Predict a random class for classification; do random ranking of users for expert retrieval.
\item \textbf{Weighted Random}: Predict randomly with probability weights based on distribution in training data for classification; for retrieval, a ranking based on answerer frequency of users on training data is used for all test samples.
\item \textbf{Text-only}: Using text CNN from \cite{kim2014convolutional} with a fully connected (FC) layer at the end.
\item \textbf{Image-only}: Using pre-trained ResNet + FC layer.
\item \textbf{Dual-net}: The model used in \cite{tamaki2018classifying}. The text representation derivation method for this model is different from the text CNN used in other models.
\item \textbf{Embedding Concatenation}: Simple concatenation of base image and text embeddings.
\item \textbf{Sum-Prod-Concat}: Element-wise sum, product, and subsequent concatenation, as done in \cite{saito2017dualnet} and \cite{tamaki2018classifying}
\item \textbf{Stacked Attention (SAN)}: Based on \cite{yang2016stacked}.
\item \textbf{Hierarchical Co-Attention (Hie-Co-Att)}: Based on \cite{lu2016hierarchical}. 
% The text representation derivation method for this model is different from the text CNN used in other models.
\item \textbf{Multimodal Compact Bilinear Pooling (MCB)}: Based on the non-attention-based mechanism in \cite{fukui2016multimodal}.
\end{itemize}
Our model (called \textbf{CQA Augmented Model}), has already been described in Section \ref{sec:final_model}.

The results for all models are presented in Table \ref{tab:results}. \textit{Random} and \textit{Weighted Random} models help to establish the difficulty of the task with respect to the performance measures used. The strong results from the \textit{Text-only} baseline indicate that for most of the samples, text contains sufficient information for both tasks, providing empirical validation for the first identified challenge in Section \ref{subsec:challenges}. Seeking improvement by combining image and text information, we get a $\sim$3\% increase by using simple embedding concatenation methods for classification, and 0.025 MRR measure increment. DualNet \cite{tamaki2018classifying} performs worse than the \textit{Text-only} model since it uses a different, less powerful text representation. Their model with our text-CNN is essentially the \textit{Sum-Prod-Concat} model. 

As expected, we don't obtain substantial improvements by using attention models, which deal better with texts that query different regions of the image. The \textit{Hie-Co-Att} model is further constructed on the premise of utilizing the image-text attention mapping at the word, phrase and sentence levels. This generalizes poorly for CQA data, where the final tasks do not benefit from learning correlation mapping between every text and image region. 

By incorporating CQA-specific augmentations, our model is able to achieve a further $>$4\% improvement on the classification task, and $>$0.015 MRR score improvement. From the perspective of being able to use the image data to improve performance, we have a substantial $\sim$8\% classification accuracy increase. 
As noted in \cite{liu2005finding}, in the expert finding task, the ground truth relevance judgment set is incomplete as there are possibly many `experts' that possess the
knowledge about a given topic, but only a small number of them actually answered the question. With this definition, comparing MRR for \textit{Text-only} and our model, for any question, the lower bound for the expected number of users to be sent a recommendation so that at-least one of them is a potential expert is down from 5 to 4.

\begin{figure*}[t]
\centering
\includegraphics[width=0.9\linewidth, height=4.2cm]{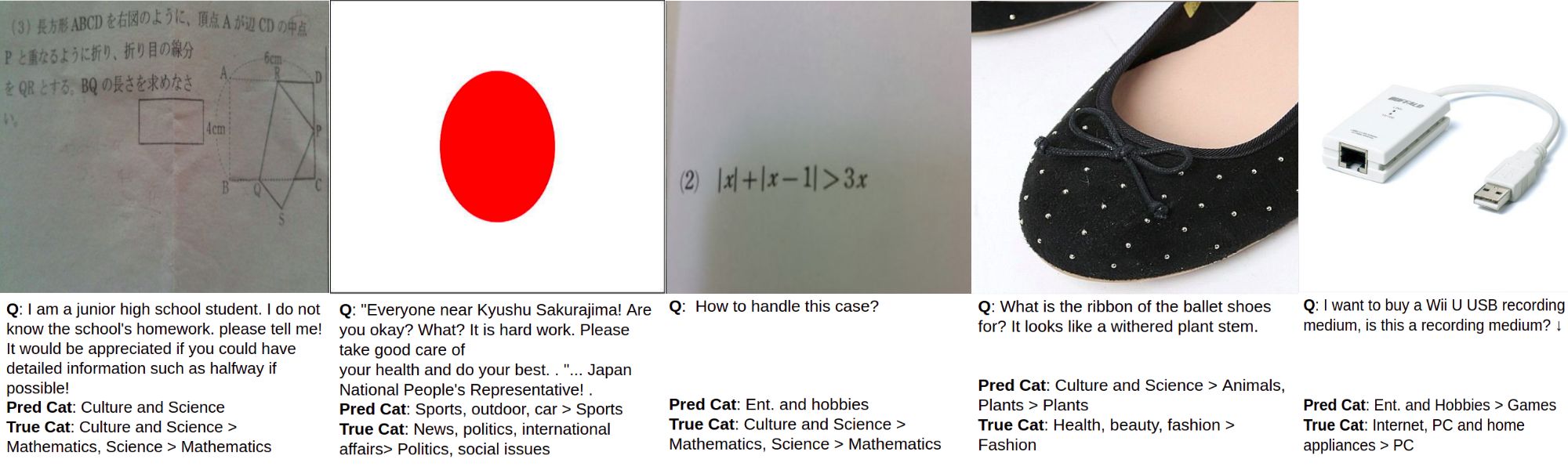}
\caption{Samples misclassified by \textit{Text-only} model, but correctly classified by
\textit{Stacked Attention Network (SAN)} model}
\label{fig:ab4}
\end{figure*}

Examples in Figure \ref{fig:ab4} help to understand the nature of samples where \textit{SAN} (taking in both image and text) performs better than the \textit{Text-only} model. Images with characteristics that are repeated across multiple samples - like mathematical problems on paper, fashion-wear items like shoes, cables \& PC equipment - are more likely to be useful towards the final task.

\begin{table}[t]
\centering
\caption{Baseline model performances vs. CQA Augmented Model on YC-CQA test split.}
\label{table:categoryclf}
\renewcommand{\arraystretch}{1.2}
\begin{tabular}
{|>{\centering\arraybackslash}m{7.5em}|>{\centering\arraybackslash}m{7em}|>{\centering\arraybackslash}m{7em}|}
\hline
\textbf{Model} & \textbf{Category Classification Accuracy (\%)}  & \textbf{Expert Retrieval: MRR} \\
\hline
\hline
Random           & 2.61        & 0.0092        \\
\hline
Weighted Random 	 & 12.16 & 0.0605 \\ 
\hline
Image-only           & 29.88          & 0.0849        \\
\hline
Text-only            & 68.32          & 0.2071       \\
\hline
Dual-net         & 68.04 & 0.2022          \\
\hline
Embedding Concatenation    & 70.52 & 0.2310 \\
\hline
Sum-Prod-Concat      & 71.35          & 0.2369 \\
\hline
SAN 1-layer        & 72.08          & 0.2375          \\
\hline
SAN 2-layer       & 72.05 & 0.2375          \\
\hline
Hie-Co-Att       & 71.87 & 0.2365         \\
\hline
MCB       & 72.01 & 0.2370         \\
\hline
CQA Augmented Model       & \textbf{76.14} & \textbf{0.2529}          \\
\hline
\end{tabular}
\label{tab:results}
\end{table}

\begin{figure*}[t]
\begin{subfigure}[b]{\textwidth}
\centering
\includegraphics[width=1.04\linewidth, height=5cm]{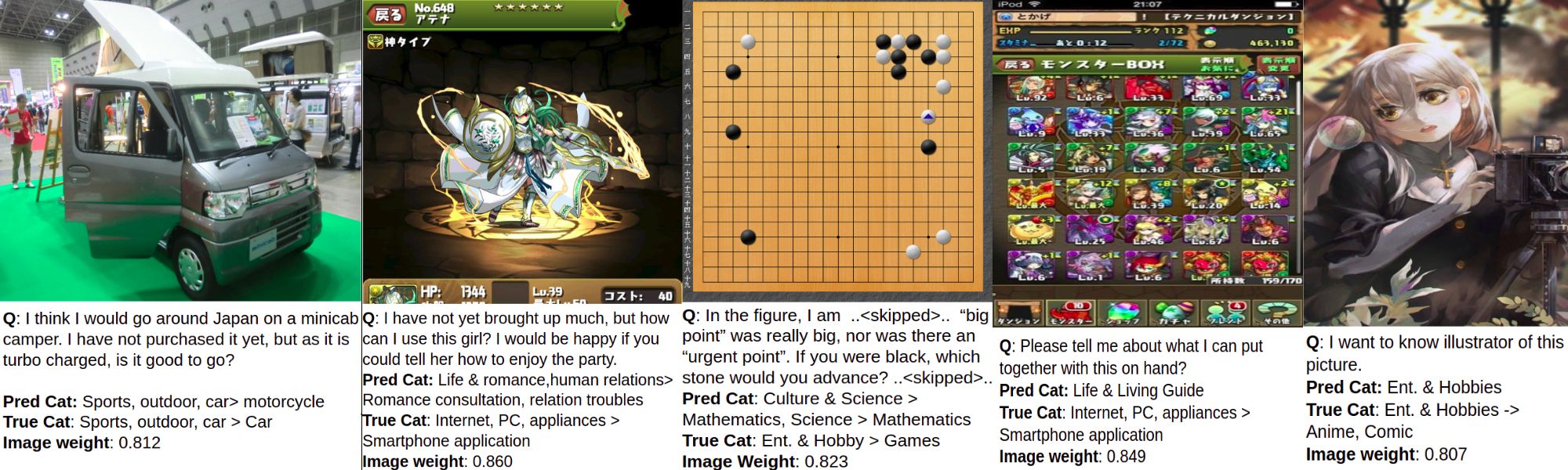}
\caption{Samples misclassified by \textit{W/O Image Weight} model, and correctly classified and assigned high global image weight by \textit{Full Model}. 
}
\label{fig:ab1}
\end{subfigure}
\begin{subfigure}[b]{\textwidth}
\centering
\includegraphics[width=1.04\linewidth, height=5cm]{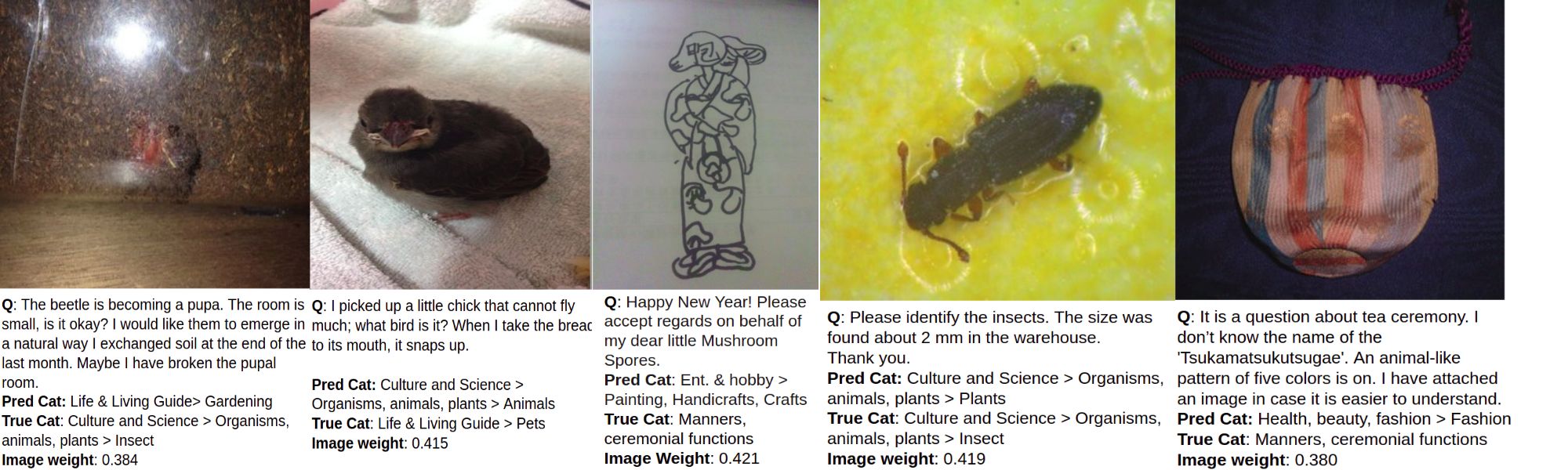}
\caption{Samples misclassified by \textit{W/O Image Weight} model, and correctly classified and assigned low global image weight by \textit{Full Model}. 
}
\label{fig:ab2}
\end{subfigure}\\\vspace{0.3cm}
\begin{subfigure}[b]{\textwidth}
\centering
\includegraphics[width=1.04\linewidth, height=5cm]{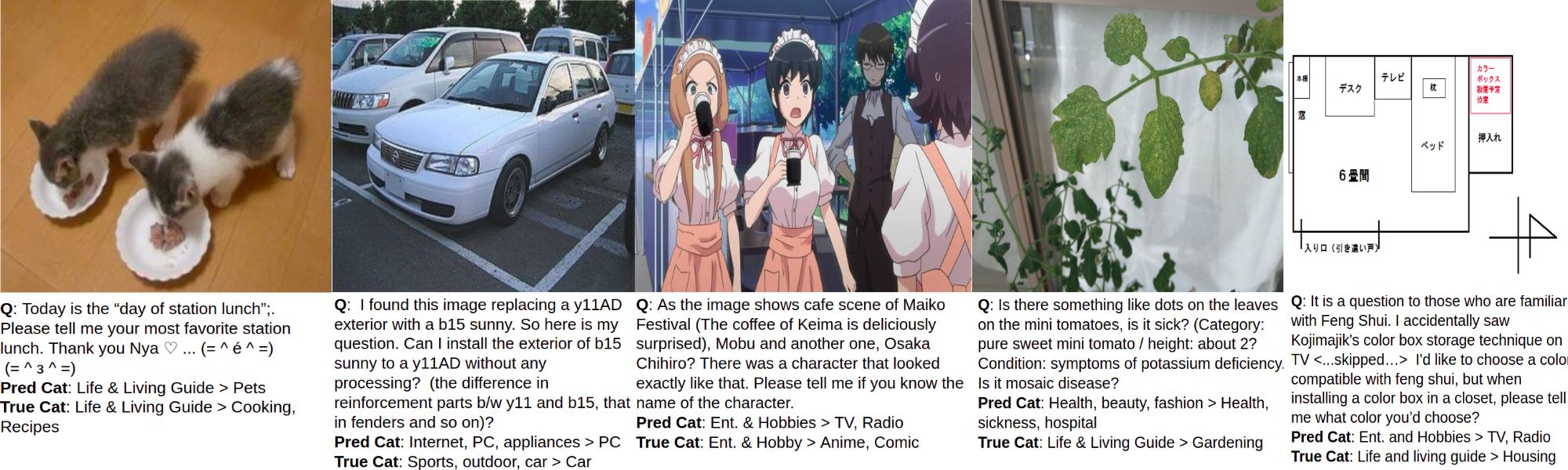}
\caption{Samples misclassified by \textit{W/O Auxiliary Tasks} model and correctly classified by the \textit{Full Model}.} 
% Automobile, plant, and animation categories occupy a big share.}
\label{fig:ab3}
\end{subfigure}

\caption{Examples misclassified by ablations (shown as `Pred Cat'), but correctly predicted by \textit{Full Model}. Image weight mentioned in (a) and (b) is the sum of attention weights for all image regions. Image weight=1 implies equal image-text contribution (similar to their role in VQA models).}
\label{fig:5}
\end{figure*}

\subsection{Ablation Analysis}
To quantify the contributions of different components in our final model, we re-evaluate after performing the following ablations:
\begin{itemize}
\item \textit{W/O image weight}: Global image weight (Section \ref{subsec:globalatt}) is removed; uses simple attention along with the auxiliary tasks. 
\item \textit{W/O auxiliary tasks}: Both auxiliary tasks (Section \ref{sec:aux}) removed. 
% Global image attention mechanism is used for the joint embedding. 
\item \textit{W/O Image-to-Texts Matching}: Among auxiliary tasks, only text-to-images matching is done.
\item \textit{W/O Text-to-Images Matching}: Among auxiliary tasks, only image-to-texts matching is done.
\item \textit{W/O Attention}: Uses global image weight without attention (Section \ref{sec:global_image_wo_attention}) instead.
\item \textit{W/O Fine-tuning}: The third step described in the training pipeline (Figure \ref{fig:4} bottom) is not performed.
\item \textit{SAN Big Att}: The text feature dimension and attention layer dimensions are increased so that the stacked attention model has similar number of trainable parameters as our full model.
\item \textit{SAN Big FC}: Two fully-connected layers are added instead.
\end{itemize}

% The results for these ablations compared to our \textit{full model} are shown in Table \ref{tab:abresults}. 
The ablation results compared to our \textit{full model} are shown in Table \ref{tab:abresults}. We can see that even with an increased parameter budget, the stacked attention network's performance doesn't improve. The dip in performance of the \textit{W/O Attention} model confirms our intuition that attention contributes better after having learned the grounding features through auxiliary tasks.

We observe significant dips in performance compared to the \textit{Full Model} when using \textit{W/O Auxiliary Tasks} or \textit{W/O Image Weight} model,  providing evidence that solutions to both of the identified challenges are crucial in improving the model.

We further investigate this by looking at randomly sampled qualitative examples presented in Figure \ref{fig:5}. We start by evaluating the contribution of the global image weight feature. Figure \ref{fig:5}(a) and \ref{fig:5}(b) present examples misclassified by \textit{W/O Image Weight} but correctly classified by \textit{Full Model} by assigning low and high global image weight ($\alpha_{m+1}$ value from Section \ref{subsec:globalatt}) respectively. The second and the fourth images in Figure \ref{fig:5}(a) come from a popular Japanese smart-phone game, with similar screenshots featuring across many samples. Other common image themes are that of automobiles and animation. Also, most of these samples' text can be judged as ambiguous for text-only based classification. This demonstrates the capability of the model to attribute more attention to the image when image features are useful and textual information is not conclusive enough.
% We believe that such image features are important when the text provides low information as in the first and the fifth examples in Figure \ref{fig:5}(a). 
On the other hand, the text features in Figure \ref{fig:5}(b) samples can be seen as strong, with difficult to interpret images - demonstrating the cases where the model succeeds by ignoring the image and focusing on text. The effect of grounding can be seen in Figure \ref{fig:5}(c), especially in the fourth example where the 
image-text combination is crucial for disambiguation over the `{\it sickness}' and `{\it gardening}' categories that can be inferred from the text.
Also, examples from the automobile and animation categories are frequently observed in these misclassified cases, where generally both image and text information provide clues for effective classification. These examples qualitatively validate the usefulness of our proposed CQA-specific solutions. 

\begin{table}[t]
\centering
\caption{Results for bigger SAN and ablations}
\label{table:categoryclf2}
\renewcommand{\arraystretch}{1.2}
\begin{tabular}
{|>{\centering\arraybackslash}m{9em}|>{\centering\arraybackslash}m{6.5em}|>{\centering\arraybackslash}m{6em}|}
\hline
\textbf{Model} & \textbf{Category Classification Accuracy (\%)}  & \textbf{Expert Retrieval: MRR} \\
\hline
\hline

SAN Big Att      & 72.48          & 0.2379 \\
\hline
SAN Big FC        & 72.37          & 0.2376          \\
\hline
W/O Image Weight           & 74.84        & 0.2499        \\
\hline
W/O Auxiliary Tasks 	 & 74.17 & 0.2474 \\ 
\hline
W/O Image-to-Texts           & 75.23          & 0.2510        \\
\hline
W/O Text-to-Images            & 75.06          & 0.2505       \\
\hline
W/O Attention    & 75.14 & 0.2504 \\
\hline
W/O Fine-tuning    & 75.82 & 0.2518 \\
\hline
Full Model       & 76.14 & 0.2529          \\
\hline
\end{tabular}
\label{tab:abresults}
\end{table}

\section{Discussion}

Testing the generalizability of our methods on other multimodal CQA platforms requires good data preparedness level for multimodal data, and ensuring that the end-tasks are of practical significance by designing them in line with the requirements of the specific platform. While this is out of the scope of this work, we are optimistic about generalizability since both of our proposed modifications are driven by multimodal data characteristics omnipresent in the  web domain - such as increased image diversity, image-text information balance, and noisy, lengthy textual component. Our model uses no hand-crafted features specific to YC-CQA, and learning is done end-to-end. While other CQA datasets are bound to bring along additional problems, our approaches promise solutions for two important ones: good joint representation learning in imbalanced information setting, and improving visual grounding.
\section{Conclusion}
We presented the challenges and solutions for dealing with classification and expert retrieval tasks on multimodal questions posted on the YC-CQA site. Among approaches at the intersection of vision and language, using representations from VQA models suits the problem best. However, upon a thorough investigation of the comparison between the two datasets, we identified two fundamental problems in direct application of VQA methods to CQA: varying image information contribution in different samples, and poor learning of grounding features. We demonstrated that our model - based on our proposed solutions of learning an additional global image weight, and better grounding features through auxiliary tasks - outperformed baseline text-only and VQA models on both tasks. 
% Both tasks are influential in improving the overall CQA experience on the website for the users, and the models are generalizable across different Q\&A platforms with growing multimodality. 
The performance on the two tasks and qualitative assessment from the ablations shows that our two proposed approaches are promising for tackling the noisy image-text query data in the web domain. Since we base our work off VQA models, it also opens interesting avenues for future research, including identifying the multimodal CQA questions that can be answered using modified versions of models developed in this study. 
% We also seek to append question-similarity information to our image-based CQA dataset, to 
% (i) provide a new question similarity dataset with the challenge of multi-modal inputs and
% (ii) 
% further test the generalization of our models for the similarity task, which is again of key practical significance for platforms with image-based questions \& discussions for duplicate content detection.

% \clearpage

\bibliographystyle{ACM-Reference-Format}
\bibliography{bibliography}

\end{document}